\documentclass[runningheads]{llncs}

\usepackage{eccv}

\usepackage{eccvabbrv}

\usepackage{graphicx}
\usepackage{booktabs}
\usepackage{bbding}
\usepackage{multirow}
\usepackage{wrapfig}
\usepackage{appendix}
\usepackage{setspace}
\usepackage{float}
\usepackage{makecell}
\usepackage[misc]{ifsym} 
\usepackage[accsupp]{axessibility}  

\renewcommand{\vec}[1]{\boldsymbol{#1}}
\newcommand{\mypara}[1]{\noindent\textbf{#1}}

\usepackage{hyperref}
\usepackage{orcidlink}

\makeatletter
\def\thanks#1{\protected@xdef\@thanks{\@thanks
        \protect\footnotetext{#1}}}
\makeatother

\begin{document}

\title{Explore the Potential of CLIP for Training-Free Open Vocabulary Semantic Segmentation}

\titlerunning{CLIPtrase}

\author{Tong Shao\orcidlink{0009-0004-5436-4122}\and
Zhuotao Tian\orcidlink{0000-0003-2698-6923}$^{\textrm{\Letter}}$\and
Hang Zhao\orcidlink{0009-0008-7252-3827}\and
Jingyong Su\orcidlink{0000-0003-3216-7027}$^{\textrm{\Letter}}$
}

\thanks{$\textrm{\Letter}$: Corresponding Authors}
\authorrunning{Shao et al.}

\institute{Harbin Institute of Technology, Shenzhen, China\\
\email{22S151082@stu.hit.edu.cn, tianzhuotao@hit.edu.cn, 200110431@stu.hit.edu.cn, sujingyong@hit.edu.cn}
}

\maketitle
\begin{abstract}

CLIP, as a vision-language model, has significantly advanced Open-Vocabulary Semantic Segmentation (OVSS) with its zero-shot capabilities. Despite its success, its application to OVSS faces challenges due to its initial image-level alignment training, which affects its performance in tasks requiring detailed local context. Our study delves into the impact of CLIP's [CLS] token on patch feature correlations, revealing a dominance of "global" patches that hinders local feature discrimination. To overcome this, we propose CLIPtrase, a novel training-free semantic segmentation strategy that enhances local feature awareness through recalibrated self-correlation among patches. This approach demonstrates notable improvements in segmentation accuracy and the ability to maintain semantic coherence across objects.
Experiments show that we are 22.3\% ahead of CLIP on average on 9 segmentation benchmarks, outperforming existing state-of-the-art training-free methods. The code are made publicly available at \url{https://github.com/leaves162/CLIPtrase}
  
\end{abstract}

\section{Introduction}
\label{sec:intro}

CLIP\cite{clip}, as a vision-language foundation model, has
gained significant popularity in recent years\cite{coop,cocoop,maple,rpo,clipes,clipood}. Its remarkable zero-shot generalization capability has played a crucial role in advancing Open-Vocabulary Semantic Segmentation (OVSS)\cite{simplebaseline,maskclip,pacl,gfsseg,lseg,lisa}.
It is often employed as an encoder, and, in order to uphold the zero-shot generalization capability for OVSS, researchers have focused on developing intricate decoder designs to accommodate the pixel-level perception\cite{zegclip,freeseg,hdmnet,zegformer,pfenet}. 

However, incorporating complex decoders to process the features extracted from the fixed CLIP model overlooks the fact that CLIP was initially trained by image-level alignment. Consequently, the globally aligned image-text features may not be well-suited for semantic segmentation that primarily relies on dense features with strong local context discrimination capabilities. 

To delve deeper into the above issue, we conduct an analysis regarding the correlations within the CLIP patch features, as shown in Figure \ref{fig:reason}. We can observe that the [CLS] token in CLIP, which is utilized for image-text alignment, may disrupt the patch correlations. Specifically, in the deeper layers, the [CLS] token does not directly attend to the patches where the target objects in the image are located. Instead, several patches in the image gradually assume a global field of view, as displayed by the bright bars in the inter-patch attention maps of layer 9 and layer 11. These patches exhibit high attention towards all other patches, effectively aggregating crucial information from the entire image. Consequently, the [CLS] token primarily focuses on these "global" patches, while the attention weights assigned to other patches are nearly negligible.

\begin{figure}[tb]
  \centering
  \includegraphics[height=4.2cm]{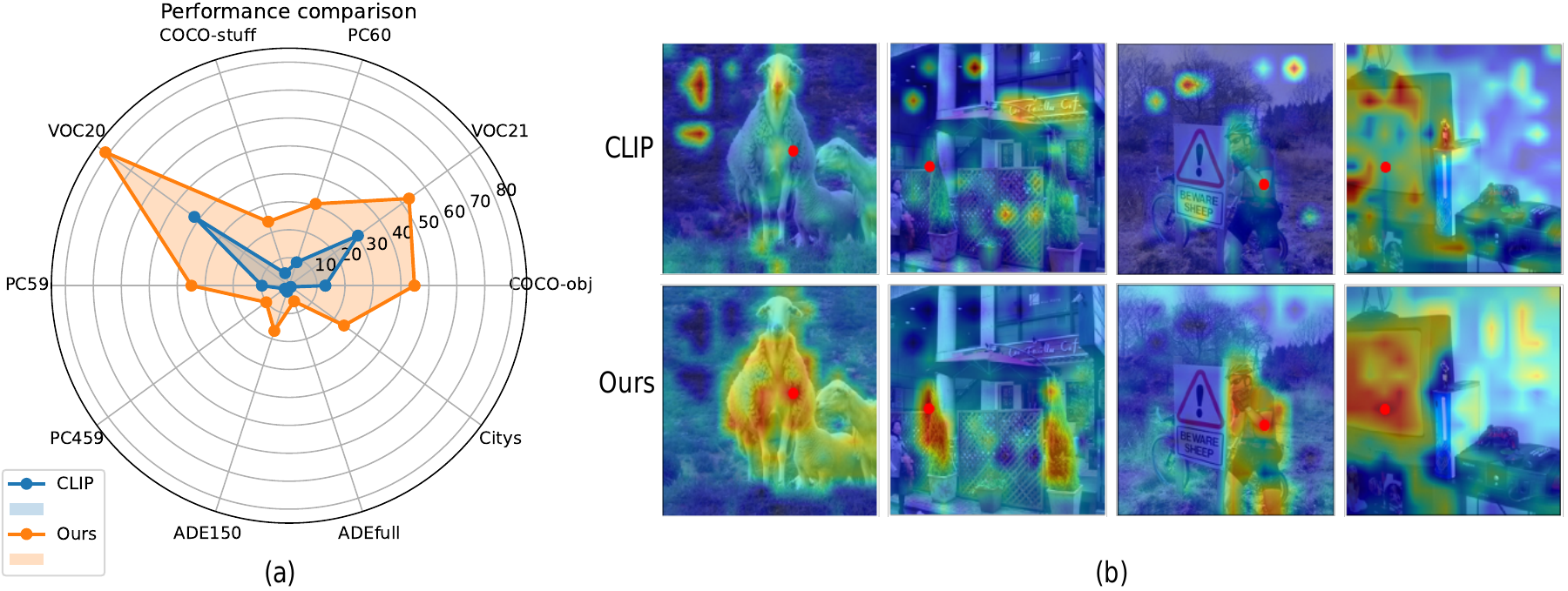}
  \caption{\textbf{Comparison between our method and CLIP.} (a): Performance comparison on open-vocabulary semantic segmentation of our model and CLIP. (b): Comparison of randomly selected patch attention response heatmaps between our model and CLIP. The red dot in the picture is the selected patch position.
  }
  \label{fig:head}
\end{figure}

The few "global" patches potentially facilitate the alignment between the entire image and text query by reducing the elements that the [CLS] token needs to concentrate on. However, it may negatively affect the local correlation. This is because the "global" patches have to gather information from all other regions to provide global cues for the [CLS] token, and disturb the correlation between local patches and the surrounding context consequently. As illustrated in the lower section of Figure \ref{fig:reason}, it can result in a decline in performance for semantic segmentation tasks. This observation raises a crucial question: \textit{Can we enhance the utilization of CLIP features by reconstructing inter-patch correlations?}

To address this, we introduce a novel \textbf{TRA}ining-free semantic \textbf{SE}gmentation approach, called \textbf{CLIPtrase}, that involves calculating self-correlation to enhance the attention of features towards their neighboring patches, consequently improving local awareness. As illustrated in Figure \ref{fig:head}(b), our enhanced CLIP enables clear discrimination between objects and backgrounds, while also revealing notable correlations among multiple objects of the same class. Subsequently, we employ a clustering and denoising process where all patches within a region compute similarity with text features and collaboratively determine the candidate classes for their respective region. As depicted in Figure \ref{fig:head}, our method exhibits significant improvements in both segmentation task performance and the visualization of patch semantic correlations, surpassing the initial CLIP model.

To summarize, our contributions are as follows:
\begin{itemize}
\item We conduct a new analysis of the limitations of CLIP in terms of semantic segmentation tasks and identify the issue brought by the "global patches".
\item To alleviate the issue, we propose a simple yet effective training-free strategy to enhance the vanilla CLIP model, termed CLIPtrase. 
\item The enhanced CLIP can be directly applied to semantic segmentation tasks, and be combined with SAM\cite{sam} to complement precise classification boundaries of SAM with semantic generalization of CLIP, achieving state-of-the-art performance in training-free OVSS. 
\end{itemize}

\section{Preliminary}
\label{sec:preliminary}
In this section, we will provide an introduction to the background of CLIP and the closely related OVSS models. More details regarding the related work are presented in the supplementary file.


\mypara{CLIP.}
Contrastive Language-Image Pre-training (CLIP~\cite{clip}) is a multi-modal foundation model. 
It achieves impressive zero-shot generalization by engaging in "image-text" contrastive learning using a vast dataset of images. 
CLIP adopts a [CLS] token within the vision transformer, similar to the text classification setting, to represent the comprehensive features of the image. This enables the alignment between image and text, with the text that bears the closest resemblance to the features of the [CLS] token being considered as the predicted class of the image:

\begin{equation}
    \vec{p} = \operatorname{argmax}(\operatorname{Sim}(\vec{t},[CLS])),
    \label{eq:clip}
\end{equation}
where $\operatorname{Sim}$ calculates the similarity between visual [CLS] token and text embedding $\vec{t}$, to obtain the prediction $\vec{p}$. 

\mypara{CLIP-based Open Vocabulary Semantic Segmentation.} 
Since CLIP is primarily trained for image-text alignment, it faces challenges in directly generating precise pixel-level predictions like segmentation~\cite{cac,decouplenet,tian2022generalized,tian2020prior}. Despite some proposed approaches\cite{zegformer,denseclip,catseg,clsclip,mvpseg} that attempt to extract detailed cues from CLIP itself for segmentation, achieving satisfactory performance remains difficult. Consequently, some approaches in the literature alternatively leverage CLIP as an encoder, extracting semantic features that auxiliary trainable modules can utilize to produce segmentation results. They can be categorized into two directions as follows:
\begin{itemize}
    \item Decoder-based: The essence of this idea is to design sophisticated decoders to refine CLIP features to adapt to semantic segmentation while retaining the generalization~\cite{deop,tcl,pmaskclip,scan,lmseg,groupvit,zeroseg}. In SAN~\cite{san}, Xu et al. propose a side network that generates masks concurrently with CLIP. These masks are subsequently employed as attention biases to capture the relevant features of each mask's corresponding [CLS] token, facilitating accurate mask classification.
    \item Fine-tuning-based: Another direction considers directly fine-tuning CLIP for segmentation~\cite{maft,pacl,lseg,ovseg}. For example, 
    MAFT~\cite{maft} uses the idea of attention bias to fine-tune the CLIP to adapt the mask classification. Additionally, OV-Seg\cite{ovseg} combines these two lines, it observes a domain shift problem when directly using CLIP 
    to classify images masked by the proposals yielded by MaskFormer~\cite{maskformer}. To tackle this issue, OV-Seg uses additional masked images to fine-tune the CLIP and make it as an encoder, so as to excel and effectively handle the specific context involving masked images.
\end{itemize}


\begin{figure}[tb]
  \centering
  \includegraphics[height=5.9cm]{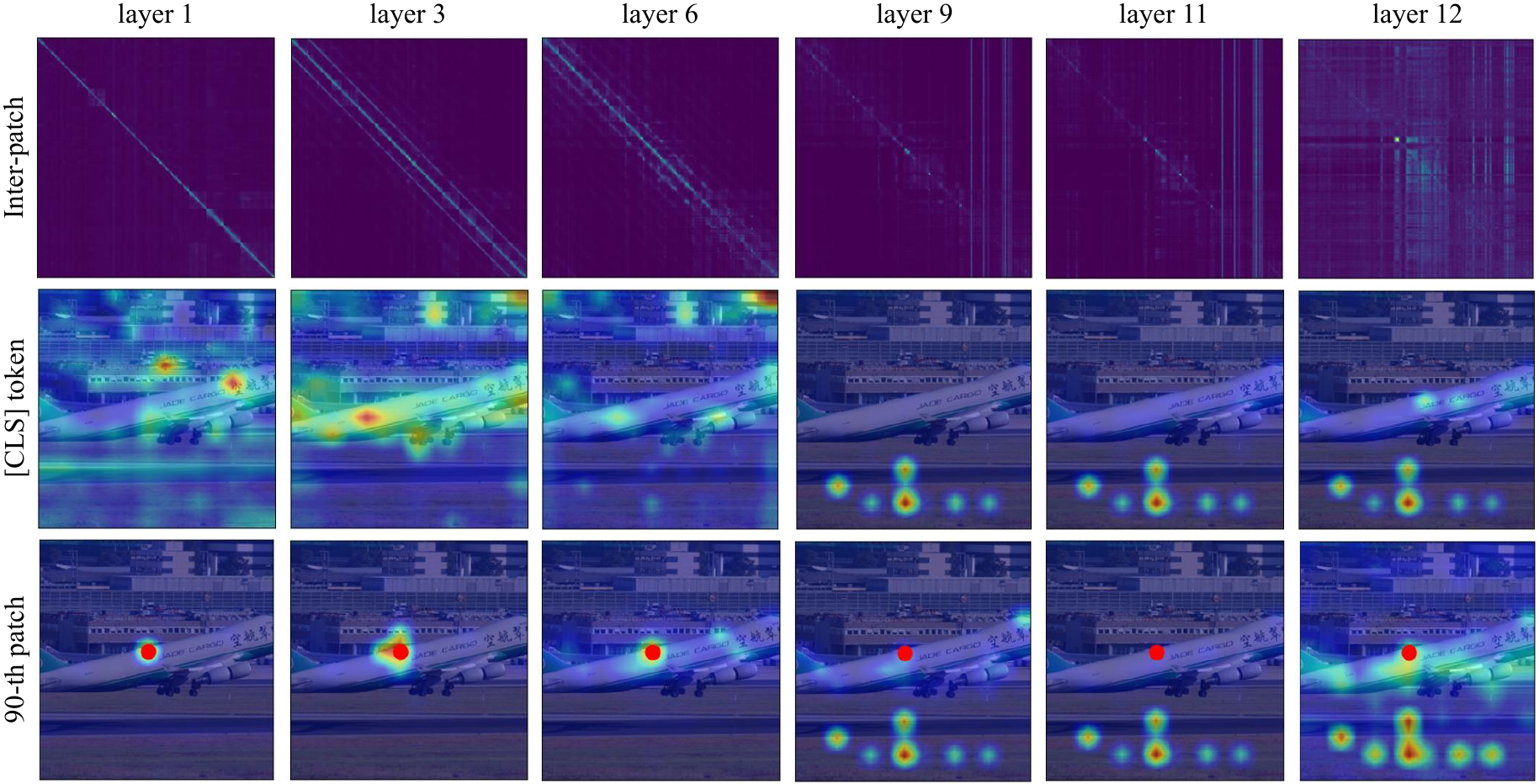}
  \caption{\textbf{Visualization of the "global" patch phenomenon in the attention map of different layers of ViT\cite{vit} in the CLIP visual branch.} Inter-patch attention map is the attention weight map between all patch features, the size is 196*196. [CLS] token attention map is the attention weight matrix of [CLS] token on all patch features, interpolates from 14*14 to 224*224, and displays in the form of a heat map. 90-th patch attention map is the attention weight of a randomly selected patch, the red dot in the image is the selected patch position. Its display method is the same as the [CLS] token. More visualization are presented in supplementary file.
  }
  \label{fig:reason}
\end{figure}

\section{Key Observations}

Despite the decent results demonstrated by the two approaches in Section \ref{sec:preliminary}, they still possess their respective issues. For the decoder-based method, the CLIP's features are obtained via a frozen CLIP model without specific adjustments to adapt them for semantic segmentation. Instead, this issue is addressed through the addition of an external decoder. On the other hand, although fine-tuning allows for a certain degree of adaptation of CLIP's image-text aligned features, it carries the risk of overfitting to specific scenarios, e.g., the domain of the masked images used for fine-tuning, leading to a decline in performance. Therefore, we need to consider whether it is possible to optimize CLIP's features without fine-tuning, in order to unearth more information that can aid in semantic segmentation. To answer this question, we start by investigating the correlations between the [CLS] token and the patch tokens.


\mypara{The "global patch".} 
The [CLS] token, used for achieving image-text alignment alongside the text embedding, provides a holistic representation of the entire image. It captures the essence or the most prominent information regarding objects within the image. 
However, since the patch tokens do not directly interact with the text embedding, their dynamics are not apparent. To delve into the intricate behaviors of the [CLS] token and the patch tokens, we examine the attention maps across different layers, as depicted in Figure \ref{fig:reason}.

Contrary to our assumptions, the [CLS] token does not directly interact with patches where target objects are located. Instead, in deeper layers, certain patches emerge as proxies for a global field of view, herein termed "global" patches. They attract high attention weights across the board, visible as bright stripes in the inter-patch attention maps of layer 9 and layer 11.
The attention map for the [CLS] token shows it mainly connects with these "global" patches, while attention to other areas is minimal. The causes of this phenomenon may be:
\begin{itemize}
    \item The high number of patches complicates the learning process for the [CLS] token. Similar to how models tend to pick up on fewer, distinct and sparse features, we think the "global" patch acts similarly, by significantly reducing the number of patches the [CLS] token needs to focus on.
    \item Much of the visual information are unnecessary for [CLS] token to accomplish the feature alignment. Visual details usually have more to offer than the text category it needs to match, especially when considering backgrounds or parts related to other co-occurring objects. Therefore, the "global" patches capture the essential information from the image by summarizing the visual content, as proxies, to provide the necessary visual essence to the [CLS] token.
\end{itemize}

Now there are studies~\cite{clstoken, register} that hold similar views to ours, which can support the credibility of our statement to a certain extent.



\mypara{The effects brought to the local patches.} 
For different local patches, as illustrated in Figure \ref{fig:reason}, it is evident that the attention weights of different local patches are predominantly influenced by the global patches. This observation reveals a lack of correlation between patches sharing the same semantics. As previously mentioned, the global patches exhibit higher attention weights towards all patches. Consequently, due to the softmax operation, the attention weights of individual local patches are primarily determined by their correlation with the global patches. This leads to a considerable suppression of the original semantic relevance between patches, rendering it nearly negligible.

\mypara{Our thoughts.} 
The learning process of the [CLS] token naturally gives rise to the prominence of the global patches. While the global patches contribute to the learning of the [CLS] token, they also significantly undermine the semantic correlation between patches. This loss of semantic correlation is particularly detrimental for dense feature tasks such as semantic segmentation. Consequently, this may be one of the primary reasons why CLIP alone is not inherently suitable for directly handling dense feature tasks.

\section{Method}
Through the analysis, we have identified the mechanisms by which the semantic correlation among patches is diminished within CLIP. This inspires us to restore the semantic correlation to enable CLIP's seamless application to semantic segmentation tasks. Considering that the primary cause of reduced semantic correlation stems from the dilution of original semantic attention, we focus on reinforcing the semantic alignment among patches, especially those with similar meanings or those in close proximity. 

To this end, we propose CLIPtrase, as a strategy to enable the direct adaptation of CLIP for semantic segmentation tasks without additional training.
CLIPtrase is composed of three core components: Semantic Correlation Recovery, Patch Clustering, and Denoising. Details are as follows.

\begin{figure}[tb]
  \centering
  \includegraphics[height=5.0cm]{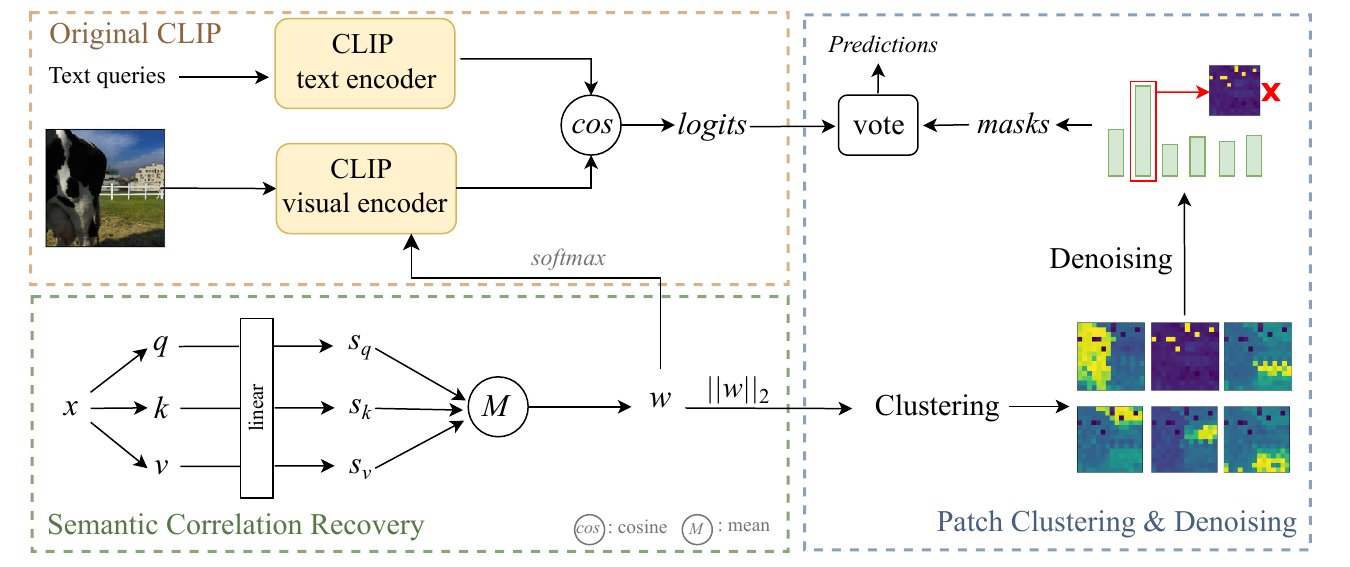}
  \caption{\textbf{Illustration of the key components for our CLIPtrase framework.} The semantic correlation operation restores the semantic location between patches. While the restored $w$ continues to forward to obtain CLIP visual features, we use clustering to obtain prototype attention weights of different categories and generate masks to improve classification results and refine object boundaries. All modules in the model are frozen to accomplish training-free setting.
  }
  \label{fig:model}
\end{figure}

\subsection{Semantic Correlation Recovery}
\label{sec:scr}

To preserve CLIP's generalization capabilities to the greatest extent, our semantic correlation restoration focuses on the final layer of the visual encoder.

We hypothesize that the input image is propagated through the shallow layers of the Vision Transformer (ViT) to yield the feature representation \(\vec{x} \in \mathbb{R}^{B \times (HW+1) \times D}\), where \(B\) represents the batch size, \(H\) and \(W\) the number of patches segmented from the image's height and width respectively, and extra one is [CLS] token. \(D\) is the dimension of the intermediate features within the CLIP visual encoder.

The multi-head attention mechanism within the original CLIP model operates as follows: the input $\vec{x}$ is first transformed by the linear layer denoted as \(\sigma(\vec{x})\). This transformed output is then divided into three distinct streams: queries (\(\vec{q}\)), keys (\(\vec{k}\)), and values (\(\vec{v}\)). Each stream undergoes processing by multiple attention heads, facilitating the model's capacity to learn diverse aspects of the data. Finally, the model applies attention weights across these streams, culminating in the aggregated output post-attention.
This process is shown as follows:
\begin{equation}
\begin{split}
    \vec{q}_{i} = \sigma_{i,q}(\vec{x}),\ \vec{k}_{i}& = \sigma_{i,k}(\vec{x}),\ \vec{v}_{i} = \sigma_{i,v}(\vec{x}),\ i\in [0,H-1]\\
    \vec{z}_{i}& = \operatorname{Softmax}(\frac{\vec{q}_i^{T}\vec{k}_i}{\sqrt{d_{k}}})\vec{v}_{i}\\
    \vec{x}& = \operatorname{Concat}(\vec{z}_{0},...,\vec{z}_{H-1})
    \label{eq:1}
\end{split}
\end{equation}
where $H$ is the number of heads, $d_{k}$ denotes the number of dimensions of $\vec{k}$, $\vec{z}$ is the intermediate output, and the operations of other layers are omitted here. 

To enhance the attention of each patch towards itself and other patches within the same semantic region, we adopt self-correlation, denoted by \(\gamma(\cdot)\). For this purpose, Cosine similarity is used to ascertain the semantic correlation across each branch of \(\vec{q}\), \(\vec{k}\), and \(\vec{v}\) along the feature dimension. Subsequently, these correlations are averaged to procure a more consistent semantic correlation matrix \(\vec{w} \in \mathbb{R}^{B \times (HW+1) \times (HW+1)}\):
\begin{equation}
\begin{split}
    \vec{q}_{i} = \sigma_{i,q}(\vec{x})&,\ \vec{k}_{i} = \sigma_{i,k}(\vec{x}),\ \vec{v}_{i} = \sigma_{i,v}(\vec{x}),\ i\in [0,H-1]\\
    \vec{w}_{i,q} = \gamma(&\vec{q}_{i},\vec{q}_{i}),\ \vec{w}_{i,k}=\gamma(\vec{k}_{i},\vec{k}_{i}),\vec{w}_{i,v}=\gamma(\vec{v}_{i},\vec{v}_{i})\\
    \vec{w}&=\frac{1}{3H}\sum_{i=0}^{H-1}(\vec{w}_{i,q}+\vec{w}_{i,k}+\vec{w}_{i,v})
\end{split}
\end{equation}


\begin{wrapfigure}{r}{0.53\textwidth}
  \includegraphics[height=2.0cm]{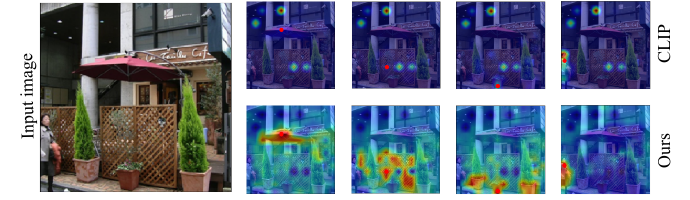}
  \caption{\textbf{CLIP attention map before and after semantic correlation recovery.} The red dot indicates the selected patch position. Our method significantly restores the correlation between adjacent or semantically similar patches.
  }
  \label{fig:demo_recovery}
\end{wrapfigure}
Besides, it is worth noting that the rationale behind adopting Cosine similarity for assessing correlations lies in that weights formed by Cosine similarity can forcibly constrain maximum attention to itself, thereby enhancing self-correlation. Comparing with the inner product, its bounded value range ensures that the resulting matrix is more amenable to subsequent clustering operation, providing a structured and constrained space for efficiently grouping semantically related patches, which shown in Section \ref{sec:ablation_study}.

Compared with the original CLIP, the attention weights are no longer concentrate on the global patches. As shown in the Figure \ref{fig:demo_recovery}, we can restore the high correlation between patches with similar semantics or in close proximity, thereby achieving the purpose of recovery.

\subsection{Patch Clustering}
\label{sec:pc}
Through the proposed recovery process, semantic correlation is enhanced, making patch features exhibit increased similarity with those sharing the same semantics. This enhancement may also improve the alignment between the visual patch and text features, as evidenced by the later experiments. 

Consequently, we take a step further by directly clustering based on enhanced semantic correlation to derive masks for individual objects, resulting in a new training-free pipeline for open-vocabulary semantic segmentation. Within each clustered region, patches can calculate their similarity with text features and collectively identify the candidate classes pertinent to their respective region.


\mypara{Clustering based on density. }
In our method, we apply the density-based clustering technique DBSCAN\cite{dbscan} to the attention weights, denoted as the function $\operatorname{DBSCAN}(\cdot)$.
The essence of density clustering lies in determining categories based on the proximity of samples. DBSCAN asserts that points within a neighborhood are considered density-reachable, and a class is defined as the largest collection of density-reachable points.

Specifically, we perform L2 normalization upon the attention weight $\vec{w}/||\vec{w}||_{2}$ for DBSCAN clustering. It is important to mention that the attention weight $\vec{w}$ does not include the [CLS] token, resulting in a shape of $\vec{w}\in \mathbb{R}^{HW\times HW}$. For clarity, we omit the batch size in subsequent expressions. Once we have obtained the clustering result $\vec{c}$ with $N$ clustering results that assigns cluster IDs for different patches, we can obtain the prototype $\vec{p}_{k}$ of the $k$-th class as follows:
\begin{equation}
    \vec{c} = \operatorname{DBSCAN}(\vec{w}/||\vec{w}||_{2})\in \mathbb{R}^{HW}
\end{equation}
\begin{equation}
    \vec{p}_{k} = \frac{1}{N_{k}}\sum_{i=1,\vec{c}_{i}=k}^{HW}\vec{w}_{i},
\end{equation}
The prototype of each class is denoted as $\vec{p}_{k}$, and $N_{k}$ represents the number of samples in the $k$-th class. Consequently, the resulting clustering attention weight prototypes can be represented as $\vec{p}\in \mathbb{R}^{N\times H\times W}$, where $N$ is the number of clustering. The attention weight matrix obtained by using Cosine similarity is not sensitive to the parameters of DBSCAN during clustering, which we will mention in subsequent Section \ref{sec:ablation_study}.

\mypara{The semantic segmentation results.}
After obtaining the prototype attention weights, we proceed to yield the semantic segmentation results $\vec{m}$ for different classes in an adaptive manner by comparing their clustering results:
\begin{equation}
    \vec{m} = \operatorname{argmax}(\vec{p}).
    \label{eq:mask}
\end{equation}
This allows the attention weights to dynamically determine the category of the object edges. By considering the logits of all pixels within each mask, a collaborative decision is made to determine the class of the corresponding region.

Specifically, we use the prediction results of CLIP to vote within each mask to obtain the final pixel-level prediction results.
As illustrated in Eq.~\eqref{eq:clip}, we use the enhanced CLIP visual encoder to obtain patch features $x\in \mathbb{R}^{HW\times D}$, and calculate the logits ($\mathbb{R}^{C\times H\times W}$) with the text features $t\in \mathbb{R}^{C\times D}$, where $C$ is the number of text classes. We omit the interpolation operation from patch to original image. Finally, the patch logits within each mask collaboratively vote for their candidate class for their respective region.



\subsection{Denoising}


\begin{figure}[!t]
    \centering
    \begin{minipage}{0.48\textwidth}
        \centering
        \includegraphics[height=4.4cm]{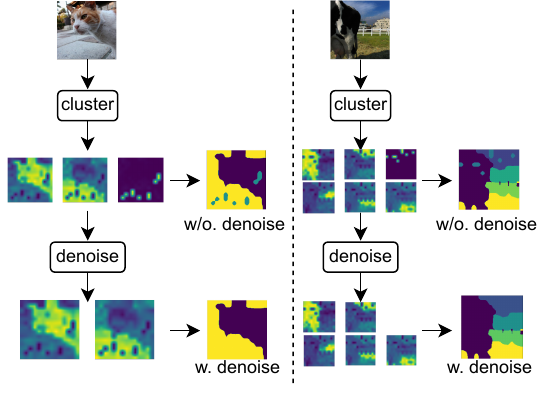} 
        \caption{\textbf{Examples of clustering and denoising process.} It can be clearly seen that there is noise caused by global patch in the results without denoising.}
        \label{fig:pc}
    \end{minipage}\hfill 
    \begin{minipage}{0.48\textwidth}
        \centering
        \includegraphics[height=2.8cm]{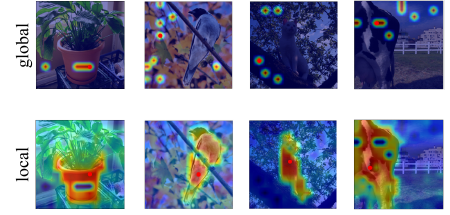} 
        \caption{\textbf{Comparison of attention maps between local and global patches.} The red dot indicates the selected patch position. Global patches have high weights between each other but they have no semantic relevance at all.}
        \label{fig:global}
    \end{minipage}
\end{figure}

Though DBSCAN can help remove certain noisy elements~\cite{dbscan}, the redundant clusters are still inevitable in the outcome, as illustrated in Figure \ref{fig:pc}, leading to inferior performance. 

\mypara{The evils in global patches.}
To investigate the cause of the noisy clusters, we re-examine the semantic correlation obtained through the recovery process discussed in Section \ref{sec:scr}. We observe that the noisy clusters are primarily formed by the global patches.
From Figure \ref{fig:global}, the reason why global patches tend to form noise clusters is due to their intrinsic cohesiveness and the high degree of similarity they share among themselves, which is not as pronounced when compared to correlation with other local image patches. 

\mypara{Identify and discard the noisy clusters.}
Based on the analysis above, we are surprised to find that for global patches after semantic correlation recovery, their attention weights with respect to other global patches are even greater than the weights with respect to themselves. On one hand, it reflects the significant disruptive effect of global patches on the correlation between local patches. On the other hand, it provides us with insights on how to identify global patches:

Specifically, we firstly extract the self-patch weights $\vec{w}_{i,i}$ from attention maps, which represent the weights between patch $\vec{i}$ and patch $\vec{i}$. Afterwards, we subtract the self-patch weights from the attention map $\vec{w}_{i}$ corresponding to patch $\vec{i}$. If $\vec{w}_{i,j}>0$, it indicates that the correlation between patch $\vec{i}$ and patch $\vec{i}$ is higher than the correlation between patch $\vec{i}$ and itself. This is clearly unreasonable, suggesting that patch $\vec{j}$ may be a global patch. Additionally, to eliminate the interference of other global patches and extreme outliers in the computation, we further calculate the average weight on patch $\vec{i}$:

\begin{equation}
\begin{split}
    \vec{w}^{*}_i = \frac{1}{HW}&\sum_{j=0}^{HW-1}\vec{w}_{i,j}-\vec{w}_{i,i}, i\in [0,HW-1]
    \\
    \vec{w}^{*} = \{&w^{*}_{i}\le0, i\in [0,HW-1]\}
\end{split}
\end{equation}
patches of which correlation between themselves are greater than those with other patches are retained, and the patches that are excluded are considered as global patches. The essence of this implies that even after semantic correlation recovery, if the attention still cannot shift from global patches to the patches themselves or surrounding regions, they will become noise in the clustering.

We utilize the refined weights $\vec{w}^{*}$, to perform clustering again and obtain the denoised clustering results $\vec{p}^*\in \mathbb{R}^{(N-1)\times H \times W}$, which means exclude the noise cluster, just like Figure \ref{fig:pc}.
Next we need to determine the masks of each object through attention weights just like Eq.~\eqref{eq:mask} :
\begin{equation}
    \vec{m}^{*} = \operatorname{argmax}(\vec{p}^*,dim=0)
\end{equation}
By using the argmax function, the magnitude of clustering results determines the allocation of masks on object boundaries. The subsequent prediction part is the same as described in Section \ref{sec:pc}.

\subsection{Integrating With Existing Works}

Our method can greatly improve the semantic location capability of CLIP, which can naturally be combined with SAM\cite{sam}. The semantic location of CLIP can be used as point prompts to obtain more accurate masks from SAM compared to the Patch Clustering module. SAM can also utilize our CLIPtrase to introduce semantic information to achieve semantic segmentation. 

It is worth noting that although SAM can replace our mask generating module, the denoising module is still necessary and is performed before SAM. This is because point prompts completely determine the segmentation area of SAM, its cost affected by noise is greater than our method based on attention weights.

In addition, our method can also be combined with other tasks, we expand this part in supplementary materials.

\section{Experiment}

\subsection{Implementation Details} 

\mypara{Experimental setting.} All our experiments are based on ViT-B/16. In order to ensure that the experiments are consistent with CLIP, image preprocessing only includes resize and normalize.
In the analysis stages such as reason analysis and ablation experiments, we uniformly use the image size of 224*224. In the main experiment, we appropriately enlarge the size to 336*336 to improve performance. 
For the hyper-parameter settings in DBSCAN clustering, we empirically set the neighborhood distance threshold $\epsilon$ to 0.7 and the neighborhood sample number threshold $min\_sample$ to 3, which are consistent for all datasets.

\mypara{Datasets.} We evaluate our method on 9 segmentation datasets, including COCO-stuff\cite{coco} with 171 categories, PASCAL VOC2012\cite{voc} with 20 categories, referred to as VOC20, PASCAL Context\cite{pc}, one with 59 categories, referred to as PC59, and another with complete 459 categories, referred to as PC459, ADE20K\cite{ade}, one is category 150, referred to as ADE150, and the other is 847 categories, referred to as ADEfull.
In addition, in order to further improve the evaluation, we further add background class prediction on COCO (take the first 80 categories), VOC20, PC59: COCO-obj, VOC21, PC60.

\mypara{Baselines.} CLIPtrase is a training-free approach, with the original CLIP as its direct baseline. Additionally, we consider related works such as CLIP~\cite{clip}, CLIP-Surgery~\cite{clipsurgery}, and SCLIP \cite{sclip}, which extensively analyze dense feature tasks. However, since SCLIP incorporates a training module, we reproduce it without training to assess the impact of a training-free approach. Furthermore, to evaluate the efficacy of our model, we select representative models requiring training in open vocabulary semantic segmentation, including GroupViT\cite{groupvit}, MaskCLIP\cite{maskclip}, TCL\cite{tcl},CAT-Seg\cite{catseg}, DeOP\cite{deop}, OVSeg\cite{ovseg}, SAN\cite{san}.

\subsection{Results}
Table \ref{tab:main} shows the comparison effect of our method with baselines. Compared with the original CLIP, our method greatly improves the potential of CLIP on semantic segmentation task. Originally, CLIP is unable to complete the semantic segmentation task, in which its performance on 6 datasets is less than 10\%. Our method can greatly improve this dilemma. 

In addition to evaluation on common benchmarks such as VOC20, we test on datasets with extremely many classes such as ADEfull and cases including background, to observe whether our method can improve semantic location on patches while ensuring category generalization as much as possible. Experiments prove that our method is effective. Our method is even close to the current SOTA training model on the ADEfull dataset (10.11\% compared with 12.6\%), which is hard even for training models. 


Our method can greatly improve the potential of CLIP to be applied to semantic segmentation task and realize training-free open vocabulary semantic segmentation.
Compared to models that require training, our method can surpass some of the baselines, but there is still a certain gap with the SOTA model like SAN. However, the biggest advantage of our model is that it can achieve training-free segmentation without training on additional datasets.

Our performance further proves that the [CLS] token in CLIP greatly damages the semantic relevance of patches.
[CLS] token only learns category difference from global patches, ignoring the semantic location between local patches.
The semantic correlation calculation restores it as much as possible, which is one of the reasons why our method is effective.
We visualize the segmentation predictions in the supplementary file.

\begin{table}[tb]
    \caption{\textbf{Evaluation results (mIoU, \%) of our method and the baseline models on ten semantic segmentation benchmarks.} The best results on each dataset in the comparison of methods that \textit{do not require training} are bolded. For more metrics (e.g., pAcc, mAcc, and fwIoU), please check the supplementary file.
  }
  \label{tab:main}
    \centering
    \scalebox{0.76}{
    \renewcommand\arraystretch{0.9}
    \begin{tabular}{c|c|ccc|cccccc|c}
    \toprule
         & \multirow{2}{*}{train}& \multicolumn{3}{c|}{w. background} & \multicolumn{6}{c|}{w/o. background} & \multirow{2}{*}{Avg.} \\
         &  & CO-obj & VOC21 & PC60 & CO-stuff & VOC20 & PC59 & PC459 & ADE150 & ADEfull  & \\ \midrule
        GroupViT\cite{groupvit} & \Checkmark & 27.5 & 52.3 & 18.7 & 15.3 & 79.7 & 23.4 & - & 10.4 & - & - \\
        MaskCLIP\cite{maskclip} & \Checkmark & 20.6 & 43.4 & 23.2 & 16.7 & 74.9 & 26.4 & - & 11.9 & - & - \\
        TCL\cite{tcl} & \Checkmark & 30.4 & 51.2 & 24.3 & 19.6 & 77.5 & 30.3 & - & 14.9 & -  & - \\
        CAT-Seg\cite{catseg} & \Checkmark & - & 78.3 & - & - & 93.7 & 57.5 & 16.6 & 27.2 & 8.4  & -\\
        DeOP\cite{deop} & \Checkmark & - & - & - & - & 91.7 & 48.8 & 9.4 & 22.9 & 7.1  & -\\
        OVSeg\cite{ovseg} & \Checkmark & - & - & - & - & 92.6 & 53.3 & 11 & 24.8 & 7.1  & - \\
        SAN\cite{san} & \Checkmark & - & - & - & - & 94 & 53.8 & 12.6 & 27.5 & 10.1  & - \\
        \hline
        CLIP\cite{clip} & \XSolidBrush & 12.63 & 17.31 & 8.91 & 4.70 & 41.06 & 9.82 & 1.88 & 2.30 & 0.79 & 11.04 \\ 
        CLIP-Surgery\cite{clipsurgery} & \XSolidBrush & - & - & - & 21.9 & - & 29.3 & - & - & - & - \\ 
        SCLIP\cite{sclip} & \XSolidBrush & 40.43 & 51.79 & 29.00 & 21.61 & 79.12 & 32.59 & 8.97 & 14.76 & 5.28 & 31.50 \\ \hline
        CLIPtrase (Ours) & \XSolidBrush & \textbf{44.84} & \textbf{53.04} & \textbf{30.79} & \textbf{24.06} & \textbf{81.20} & \textbf{34.92} & \textbf{9.95} & \textbf{17.04} & \textbf{5.89} & \textbf{33.53} \\
    \bottomrule
    \end{tabular}}
\end{table}

\subsection{Ablation Study}
\label{sec:ablation_study}
We uniformly use the image size of 224*224 which is same as CLIP and perform on four common benchmarks COCO-stuff, VOC20, PC59, and ADE150.

\mypara{Module effectiveness.} Our method consists of three modules: semantic correlation recovery (SCR), patch clustering (PC) and denoising (D).
As shown in Table \ref{tab:ablation}, we gradually add each module to show the contribution of them to the overall performance.
In addition, we analyze the time overhead by flops. 

From the results, we can see that the most contribution of our method is semantic correlation recovery, which effectively alleviates the problem of global patches and greatly enhances the semantic correlation between patches.
The semantic correlation clustering mainly uses the value of the attention weights of the clustered objects to adaptively obtain the masks of the corresponding objects, making the pixel classification consistency on the objects stronger and avoiding fragmented prediction results. 

In addition, the time overhead of SCR and D has almost no increase, it mainly consumes time during clustering.
It should be noted that this time overhead is obtained by running single-threaded DBSCAN on the CPU. If parallelization or GPU acceleration is performed, the overhead will be further reduced.

\begin{table}[tb]
    \caption{\textbf{Ablation results (mIoU, \%) of common four datasets.} The best results on average in the comparison are bolded. When comparing different methods of a specific module, other modules are kept unchanged.
  }
  \label{tab:ablation}
    \centering
    \scalebox{0.76}{
    \renewcommand\arraystretch{1.3}
    \begin{tabular}{c|cccc|cc|cc|cc|cccc}
    \toprule
     & \multicolumn{4}{c|}{Ablation} & \multicolumn{2}{c|}{Correlation} & \multicolumn{2}{c|}{Cluster} & \multicolumn{2}{c|}{Image size} & \multicolumn{4}{c}{Layers}\\
     \midrule
     Module & CLIP & +SCR & +PC & +D & Multi & COS & $f$ & $p$  & 224 & 336 & 9 & 10 & 11 & 12 \\ \hline
     COCO-stuff & 4.70 & 21.82 & 22.46 & 22.84 & 12.64 & 22.84 & 20.44 & 22.84 & 22.84 & 24.06 & 0.47 & 1.50 & 7.75 & 22.84 \\
     VOC20 & 41.06 & 77.58 & 80.43 & 80.95 & 63.55 & 80.95 & 74.58 & 80.95 & 80.95 & 81.20 & 13.65 & 11.83 & 40.83 & 80.95 \\
     PC59 & 9.82 & 32.55 & 33.21 & 33.83 & 19.94 & 33.83 & 32.42 & 33.83 & 33.83 & 34.92 & 1.75 & 4.94 & 15.78 & 33.83 \\
     ADE150 & 2.30 & 15.79 & 16.00 & 16.35 & 7.49 & 16.35 & 15.01 & 16.35 & 16.35 & 17.04 & 0.41 & 1.74 & 4.08 & 16.35 \\ \hline
     Avg. & 14.47 & 36.94 & 38.02 & \textbf{38.50} & 25.91 & \textbf{38.50} & 35.61 & \textbf{38.50} & 38.50 & \textbf{39.31} & 4.07 & 5.00 & 17.11 & \textbf{38.50} \\ \hline
     Flops & x1 & x1.2 & x1.9 & x2.0 & - & - & - & - & - & - & - & - & - & - \\
    \bottomrule
    \end{tabular}}
\end{table}

\mypara{Semantic relevance calculation.}
In Section \ref{sec:scr}, we mentioned that the semantic similarity matrix obtained by Cosine similarity has a better effect on subsequent clustering due to its determined value range.
Here we compare the performance of clustering using semantic similarity matrices obtained by inner product (multi) and Cosine (cos). From Table \ref{tab:ablation}, it can be clearly seen that the semantic similarity matrix obtained by cos in the fixed value range works very well. More results can be referred in supplementary materials.


\begin{wrapfigure}{r}{0.5\textwidth}
  \includegraphics[height=4.6cm]{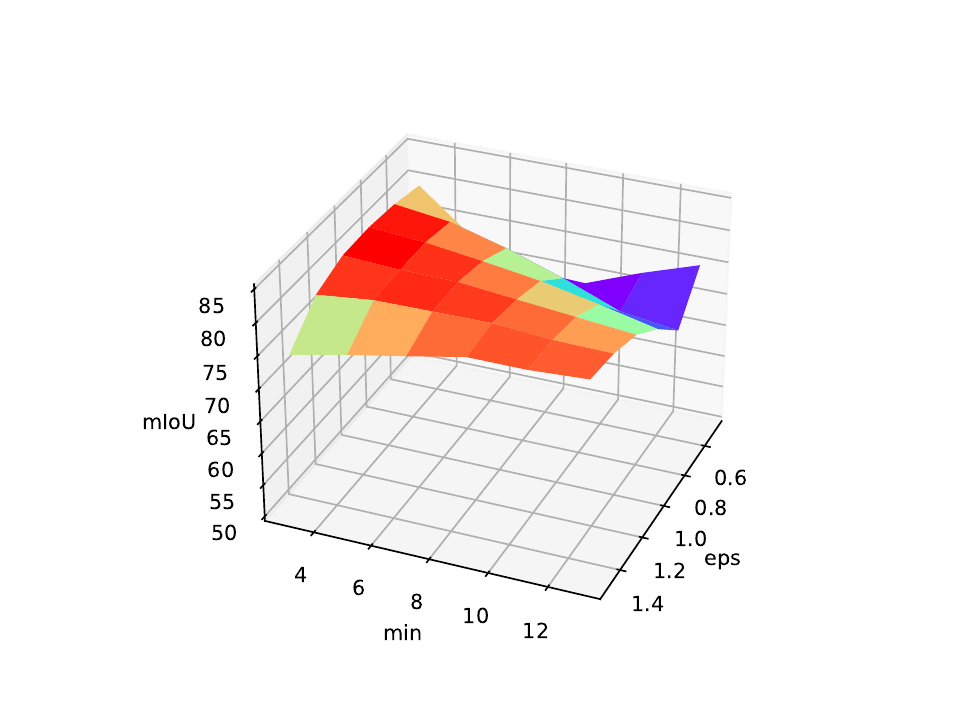}
  \caption{\textbf{Sensitivity of clustering parameter space of VOC20.} Where eps represents $\epsilon$ and min is $min\_sample$. Different colors represent different performance intervals. Except for some marginal extreme hyper-parameter values, the performance of most of the parameter space is above 75, and the best one is 81.01.
  }
  \label{fig:params}
\end{wrapfigure}
\mypara{Cluster object.}
In addition to clustering attention weights, we try to cluster features with enhanced self-correlation, which represented by $\vec{f}$, and the results are shown in the Table \ref{tab:ablation}. Since the feature dimension is higher, clustering is more difficult, so its results are lower than the attention weight prototype $\vec{p}$ clustering.

\mypara{Hyper-parameters.} They mainly contains: the image size, the clustering parameters, and select which layer to restore semantic correlation.

For the image size, we compare the performance on 224 and 336. As the size increases, the model performance improves to a certain extent, but the number of patches also increases, thereby increasing the time overhead, which requires a trade-off. For better results, we select 336 resolution in our main results. During the analysis and discussion stage, we use 224 resolution to improve efficiency.

For DBSCAN clustering, its main hyper-parameters are neighborhood distance threshold $\vec{\epsilon}$ and the neighborhood sample number threshold $min\_sample$. 
Thanks to the deterministic value range of Cosine similarity, the clustering $||w||_{2}$ jitter is small, allowing us to use a fixed set of parameters to deal with all datasets. In order to illustrate the sensitivity of our method to parameter selection, we use different permutations and combinations of $\epsilon$ and $min\_sample$ in clustering to observe fluctuations in the results of VOC20. Except for extreme values, our method can maintain stability in most parameter spaces.

As for which layer should be used to restore semantic relevance, our aim is to restore semantic correlation without destroying the CLIP modal alignment as much as possible. We try and compare the deep layers of the model in Table \ref{tab:ablation} and finally find that the last layer has the best effect.





\begin{table}[tb]
    \caption{\textbf{Evaluation results (mIoU, \%) of our method combined with SAM.} We replace the patch clustering module with SAM and advance the denoising module to weaken the impact of noise on SAM. The best results on each dataset are bolded.
  }
  \label{tab:sam}
    \centering
    \scalebox{0.8}{
    \renewcommand\arraystretch{1.3}
    \begin{tabular}{c|ccc|cccccc|c}
    \toprule
        & \multicolumn{3}{c|}{w. background} & \multicolumn{6}{c|}{w/o. background} & \multirow{2}{*}{Avg.} \\
        & CO-obj & VOC21 & PC60 & CO-stuff & VOC20 & PC59 & PC459 & ADE150 & ADEfull  & \\ \midrule
        CLIP & 12.63 & 17.31 & 8.91 & 4.70 & 41.06 & 9.82 & 1.88 & 2.30 & 0.79  & 10.02 \\ 
        SCR+PC+D & \textbf{44.84} & 53.04 & 30.79 & 24.06 & 81.20 & 34.92 & 9.95 & 17.04 & 5.89  & 33.54 \\
        SCR+D+SAM & 44.23 & \textbf{57.06} & \textbf{31.97} & \textbf{24.75} & \textbf{82.25} & \textbf{36.44} & \textbf{10.58} & \textbf{17.22} & \textbf{5.97} & \textbf{34.49} \\
    \bottomrule
    \end{tabular}}
\end{table}

\mypara{Combined with SAM.} Our method can also be combined with SAM\cite{sam} to provide semantic location. The high-resolution boundaries of SAM can further improve CLIP performance on segmentation tasks.

Here we use SAM to replace our mask generating module.
We utilize the clustered patch tags to directly interpolate the patch positions of different categories back to the original image as point prompts, and obtain more accurate masks for each object through SAM. The prompts obtained by different masks serve as background points for each other. Essentially, SAM, like the clustering module, provides refined object boundaries for our method.
As shown in the Table \ref{tab:sam}, its refined mask boundaries can further improve the performance of our method.

\section{Concluding Remarks}
\mypara{Summary.}
This paper presents an analysis of the limitations that prevent CLIP from effectively performing dense feature tasks. We observe that the global patches disrupt the semantic correlation between image patches, and propose a method to address this issue. By adopting this method, CLIP can be directly applied to segmentation tasks. Extensive experiments demonstrate that our training-free open-vocabulary semantic segmentation approach can yield considerable improvements to other training-free counterparts.

\mypara{Limitations.}
Our research delves into understanding and enhancing the capabilities of CLIP for segmentation without training. However, it's important to acknowledge that despite these improvements, there remains a discernible gap compared to the performance of state-of-the-art trainable models. Moreover, extending our analysis and enhancement techniques to tasks requiring denser feature extraction, like object detection and panoptic segmentation, may be a promising future direction.

\section*{Acknowledgements}
This work was supported by National Natural Science Foundation of China (grant No. 62376068, grant No. 62350710797), by Guangdong Basic and Applied Basic Research Foundation (grant No. 2023B1515120065),  by Guangdong S\&T programme (grant No. 2023A0505050109), by Shenzhen Science and Technology Innovation Program (grant No. JCYJ20220818102414031).

%
%

\appendix
\setstretch{1.0}
\section*{Supplementary File}
This is the supplementary file for our submission titled \textit{Explore the Potential of CLIP for Training-Free Open Vocabulary Semantic Segmentation}. This material supplements the main paper with the following content:

\setstretch{1.3}
\begin{itemize}
  \renewcommand{\labelitemi}{$\bullet$}
  \item \ref{a}. \textbf{Details of Our Method CLIPtrase}
  \begin{itemize}
    \item[--] \ref{a1}. More Explanation of Global Patch
    \item[--] \ref{a2}. Semantic Correlation Computing
    \item[--] \ref{a3}. Mask Classification
  \end{itemize}
  \item \ref{b}. \textbf{More Experiments}
  \begin{itemize}
    \item[--] \ref{b1}. Experiment Details
    \item[--] \ref{b2}. More Experiment Results
    \item[--] \ref{b3}. More applications
  \end{itemize}
  \item \ref{c}. \textbf{Related Work}
  \begin{itemize}
    \item[--] \ref{c1}. Contrastive Language-Image Pre-training
    \item[--] \ref{c2}. Open-Vocabulary Semantic Segmentation
  \end{itemize}
  \item \ref{d}. \textbf{Visualization}
\end{itemize}

\setstretch{1.0}
\section{Details of Our Method CLIPtrase}
\label{a}
This section primarily supplements some method details and consists of the following three parts: 1). further analysis of global patches, 2). efficient variants for effective semantic correlation recovery calculation, and 3). reasons for using mask classification.

\subsection{More Explanation of Global Patch}
\label{a1}
\mypara{Observation. }
As mentioned in the main text, the [CLS] token does not directly weight all patches of the entire image through an attention weight map to obtain the final result as we might imagine. Instead, the [CLS] token interacts with the global patches to serve as a stepping stone for completing the process.
Global patches: When the [CLS] token extracts overall information from the image, there are several global patches that have high attention weights for any local patch in the image, providing a global view. The [CLS] token does not directly interact with all patches but focuses on these global patches to obtain image-level information, while disregarding the weights of other local patches.

Furthermore, through our observations, we find that this phenomenon is commonly observed from the 6-th to the 11-th layers of the CLIP ViT, while with some alleviation in the final layer strangely. 

\mypara{Analysis. }
Since the CLIP model is trained solely on a contrastive learning objective function and there are no additional constraints on the [CLS] token, we believe that this phenomenon is spontaneously formed by the model during the training process. We explore why CLIP spontaneously forms global patches by considering the benefits they provide to the CLIP. By starting from the advantages offered by such global patches, we can gain insights into the reasons behind their spontaneous formation in the model.

We think there are two main reasons:\\
First of all, one of the reasons is the redundancy of image information. From the perspective of information content, even if they describe the same object, the information content of images is much higher than that of text. And in the image, in addition to the main object described, most areas are other objects and background, which are redundant for text category features. This aspect causes [CLS] token to face a lot of redundant information when extracting image-level feature representation.

The emergence of global patches can effectively alleviate this phenomenon. The global patch itself can act as a filter, allowing the [CLS] token to learn features that better match the texts. 

This can be seen from the semantic correlation visualization of images in Section \ref{d}. Often, for images with large-area backgrounds, such as large-area sky, ocean, etc., the more global patches there are. This is similar to the conclusion in the StreamingLLM\cite{clstoken} model: Xiao et al. believe that the first token in GPT acts as a "trash can", causing redundant information to be placed in this token. Our global patch here also plays a similar role and helps filter redundant information.

The second point is due to the training trend of the model. Just like general models tend to learn sparse features, CLIP's [CLS] token also has this tendency when faced with complex patch tokens. A large number of patch tokens increases the difficulty of learning of [CLS] token. The emergence of global patches can make the patch interaction process also show a sparse trend, and [CLS] token only interacts with global patches. This greatly reduces the learning difficulty of [CLS] token.

\mypara{Effects. }
Although the emergence of this global patch has several benefits we mentioned above, it also greatly destroys the semantic correlation between local patches.

Global patches maintain high weighted attention with all local patches. However, due to the existence of softmax, the high weights with the global patch gradually squeeze the correlation weights between the original patches, causing the local patches to lose their ability to pay attention to adjacent or same semantic patches.

In the final layer of CLIP-ViT, due to the objective function aligning the [CLS] token with the text, certain semantic information is fed back from the [CLS] token to the patches across the entire image, leading to a partial recovery of semantic relevance among local patches. However, due to the influence of the global patches mentioned earlier, this recovery is limited and cannot fully counteract the suppression of original semantic relevance by the global patches in the attention matrix. 

Therefore, the presence of global patches leads to a lack of semantic relevance among local patches, which is the main reason why CLIP is not well-suited for dense feature tasks such as semantic segmentation.

\subsection{Semantic Correlation Computing}
\label{a2}
In the original computation of semantic correlation recovery, we perform self-correlation separately for each branch and each head. We then calculate the mean to obtain a more stable measure of semantic relevance:
\begin{equation}
    \vec{q}_{i} = \sigma_{i,q}(\vec{x}),\ \vec{k}_{i} = \sigma_{i,k}(\vec{x}),\ \vec{v}_{i} = \sigma_{i,v}(\vec{x}),\ i\in [0,H-1]
\end{equation}
\begin{equation}
    \vec{w}_{i,q} = \gamma(\vec{q}_{i},\vec{q}_{i}),\ \vec{w}_{i,k}=\gamma(\vec{k}_{i},\vec{k}_{i}),\vec{w}_{i,v}=\gamma(\vec{v}_{i},\vec{v}_{i})
\end{equation}
\begin{equation}
    \vec{w}=\frac{1}{3H}\sum_{i=0}^{H-1}(\vec{w}_{i,q}+\vec{w}_{i,k}+\vec{w}_{i,v})
\end{equation}

where $\sigma,\ \gamma$ is the linear layer and semantic correlation we mentioned in the main text, respectively.
However, performing separate computations on each head can reduce computational efficiency. Therefore, a simple approach to streamline this calculation process is to concatenate the outputs from each head:
\begin{equation}
    \vec{q}_{i} = \sigma_{i,q}(\vec{x}),\ \vec{k}_{i} = \sigma_{i,k}(\vec{x}),\ \vec{v}_{i} = \sigma_{i,v}(\vec{x}),\ i\in [0,H-1]
\end{equation}
\begin{equation}
    q=\phi(\vec{q}_{0},...,\vec{q}_{H-1}),\ k=\phi(\vec{k}_{0},...,\vec{k}_{H-1}),\ v=\phi(\vec{v}_{0},...,\vec{v}_{H-1})
\end{equation}
\begin{equation}
    w=\frac{1}{3}(\gamma(q,q)+\gamma(k,k)+\gamma(v,v))
\end{equation}

where $\phi$ is the operation of concatenation.  Such a replacement solution reduces the computational burden and has almost no impact on performance. However, when explaining ideas, we prefer to use the calculation method in the main text because it is more intuitive.

\mypara{Rationale of semantic correlation recovery.}
In Table \ref{table:scr}, we gradually refine the scope of attention to estimate an increase in the degree of semantic correlation restoration (SCR): 1): without SCR; 2): self attention: focus on own patches, 3): normal attention: by inner product or Cosine, 4):attention within ground truth regions.
It is evident that global patches significantly disrupt semantic correlation from Figure 1,2,4 of the paper. As the region of SCR becomes more precise, the segmentation performance gradually improves in Table \ref{table:scr}, which indicates a diminishing influence of global patch.
\textit{We demonstrate that as the degree of SCR increases, the segmentation performance improves. This serves as the proof that global patch is one of the reasons causing limited segmentation.}\\
Furthermore, both Cosine similarity and inner product yield similar results from row 3,4 in the Table \ref{table:scr}. However, the unbounded range from inner product has a negative impact (lines 209-215, 387-392 in the paper) on subsequent clustering module, which shown in last two rows of Table \ref{table:scr}. Therefore, we select the Cosine similarity to implement the SCR.

\begin{table}[tb]
  \centering
  \caption{\textbf{SCR rationality validation.}}
  \setlength{\tabcolsep}{1.6mm}{
  \begin{tabular}{c|cccc|c}
    \toprule
   input=224, mIoU & VOC & ADE & COCO & PC & Avg.\\
   \hline
   w/o. SCR & 67.77 & 3.01 & 5.40 & 10.85 & 21.76 \\
   $<x_i, x_i>$ & 62.69 & 11.55 & 16.40 & 25.11 & 28.94 \\
   $<x_i, x_j>$ & 75.61 & 15.58 & 21.39 & 32.13 & 36.18 \\
   $COS(x_i, x_j)$ & 77.58 & 15.79 & 21.82 & 32.55 & 36.94 \\
   $COS(x_i, x_j), j\in GT$ & 77.71 & 16.50 & 22.57 & 33.89 & 37.67 \\ \hline
   CLIPtrase (inner product SCR) & 63.55 & 7.49 & 12.64 & 19.94 & 25.91 \\ 
   \textbf{CLIPtrase (Cosine SCR)(Ours)} & 80.95 & 16.35 & 22.84 & 33.83 & \textbf{38.50} \\ \bottomrule
  \end{tabular}}
  \label{table:scr}
\end{table}

\subsection{Mask Classification}
\label{a3}
In the main paper, our clustering design is mainly to adaptively obtain the mask corresponding to each object, and then let all pixels in the mask jointly determine the category of this area. So there may be a question here: \textit{why not do patch-text prediction directly?}

We explain in detail the reasons for this operation:

First, such a clustering operation can form masks of the same semantic areas through the common features of the semantic similarity matrix. This idea of mask and joint decision-making make most of the correct predictions in this semantic area to improve some noise predictions. Through the performance and ablation experiments in the main paper we are able to prove the effectiveness of this approach and can effectively improve the prediction noise in the image.

In addition, this approach is closer to the current methods that requires training based on MaskFormer\cite{maskformer}, which lays the foundation for us to apply it to the models those require training.

In fact, a more consistent approach with mask-based training models such as \cite{san,deop} is to use the mask obtained by clustering to perform mask pooling on the CLIP features, and use the masks as the attention bias to generate its corresponding [CLS] token for each mask. However, since this method requires separately designing and training some layers for attention bias, which violates our original intention of directly adapting CLIP to the semantic segmentation task, it was abandoned.

In addition, to further demonstrate the effectiveness of our self-correlation and the rationale behind using masks, we compare the results obtained by using DINO~\cite{dino} for masking.
Table \ref{table:dino} presents a detailed performance comparison, and it can be observed that our method exhibits similar performance to the DINO model during the masking process after undergoing self-correlation recovery.

\begin{table}[tb]
    \caption{\textbf{Our approach combined with MaskCLIP or DINOv2.}
  }
  \label{tab:main224}
    \centering
    \setlength{\tabcolsep}{2mm}{
    \begin{tabular}{c|cccc|c}
    \toprule
    input=224, mIoU & VOC & ADE & COCO & PC & Avg.\\
   \hline
   Ours + DINOv2 & 80.23 & 15.79 & 21.82 & 32.55 & 37.59\\ \hline
   \textbf{CLIPtrase(Ours)} & 80.95 & 16.35 & 22.84 & 33.83 & \textbf{38.50}\\
    \bottomrule
    \end{tabular}}
    \label{table:dino}
\end{table}

\section{More Experiments}
\label{b}
\subsection{Experiment Details}
\label{b1}
In this section, we present further details and configurations utilized in our experiments.

\mypara{Environment. }The environment we use is: CUDA version: 11.3, PyTorch: 1.12.1, GPU: NVIDIA RTX 3090*1, CLIP: CLIP-B/16, local implementation. 

\mypara{Data Proprocessing. }The data preprocessing and data enhancement solutions in this paper are consistent with CLIP preprocessing and do not add any additional operations. We maintain the order of operations of resize, crop, and normalize. The mean and variance of normalizing on the image are:[0.48, 0.46, 0.41], [0.27, 0.26, 0.28] (two decimal places).

\mypara{Hyper-parameters. }Our image sizes in experiments are 224, 336. In the subsequent clustering of the attention weights, $\epsilon=0.7,\ min\_sample=3$. The text prompts are consistent with the official implementation of CLIP. The average of 80 short sentence prompts is taken to represent the final text feature.

\begin{table}[tb]
    \caption{\textbf{Performance comparison of four evaluation indicators for image size 224.} The best average performance under each metric is bolded.
  }
  \label{tab:main224}
    \centering
    \scalebox{0.84}{
    \renewcommand\arraystretch{1.3}
    \begin{tabular}{c|cccc|cccc|cccc}
    \toprule
        Image size=224 & \multicolumn{4}{c|}{CLIP} & \multicolumn{4}{c|}{SCLIP} & \multicolumn{4}{c}{CLIPtrase (Ours)} \\
        Evaluation & pAcc & mAcc & fwIoU & mIoU & pAcc & mAcc & fwIoU & mIoU  & pAcc & mAcc & fwIoU & mIoU \\ \midrule
        COCO-obj & 23.45 & 20.58 & 14.39 & 12.98 & 49.05 & 59.43 & 37.38 & 40.68 & 50.08 & 62.50 & 38.19 & 43.56 \\
        VOC21 & 42.21 & 47.21 & 30.47 & 17.54 & 77.52 & 83.27 & 66.46 & 49.54 & 78.63 & 84.11 & 67.67 & 50.88 \\ 
        PC60 & 21.84 & 21.22 & 11.02 & 8.80 & 49.24 & 51.55 & 35.04 & 28.09 & 52.14 & 56.08 & 37.61 & 29.87 \\ \hline
        COCO-stuff & 10.85 & 12.02 & 5.66 & 4.68 & 35.24 & 40.02 & 24.42 & 21.00 & 38.90 & 44.47 & 26.87 & 22.84 \\
        VOC20 & 58.50 & 58.35 & 44.02 & 41.88 & 87.51 & 89.92 & 79.21 & 77.58 & 89.68 & 91.40 & 82.49 & 80.95 \\ 
        PC59 & 24.60 & 21.78 & 13.29 & 9.70 & 55.69 & 52.58 & 42.13 & 31.58 & 58.94 & 57.08 & 45.28 & 33.83 \\ 
        PC459 & 16.94 & 4.02 & 9.31 & 1.92 & 41.15 & 18.63 & 32.72 & 8.48 & 44.18 & 21.53 & 35.22 & 9.36 \\
        ADE150 & 5.20 & 7.61 & 2.59 & 2.25 & 33.00 & 34.43 & 23.44 & 14.46 &  38.57 & 39.17 & 27.96 & 16.35 \\ 
        ADEfull & 2.69 & 3.13 & 1.21 & 0.76 & 20.06 & 16.46 & 14.55 & 5.43 & 25.45 & 18.78 & 18.99 & 6.31 \\ \hline
        AVG. & 22.92 & 21.77 & 14.66 & 11.17 & 49.83 & 49.59 & 39.48 & 30.76 & \textbf{52.95} & \textbf{52.79} & \textbf{42.25} & \textbf{32.66} \\
    \bottomrule
    \end{tabular}}
\end{table}

\mypara{Datasets. }We have a total of 9 benchmarks in the experiments, involving 4 datasets:\\
\begin{itemize}
    \item \textbf{COCO}\cite{coco}: There are a total of 80 object classes and 91 stuff classes. We use a total of 171 classes as COCO-stuff, and a total of 81 classes of objects and additional background classes as COCO-object.
    \item \textbf{PASCAL CONTEXT}\cite{pc}: There are 459 categories in total, and we select 59 common categories as PC59, and all categories as PC459. In addition, we add an additional background as PC60 based on PC59.
    \item \textbf{PASCAL VOC2012}\cite{voc}: There are 20 categories in total, which we refer to as VOC20. In addition, we add common background categories as VOC21 based on VOC20.
    \item \textbf{ADE20K}\cite{ade}: There are 847 classes in total, and we select 150 of them as ADE150, and all classes as ADEfull.
\end{itemize}

\mypara{Evaluation Protocol. }Following the common practice\cite{maskformer,openseg,simplebaseline}, we use the mean of class-wise intersection over union (mIoU) to measure the performance.
In addition, we also report on the performance of mean accuracy (mAcc), pixel accuracy (pAcc), and frequency weighted intersection over union (fwIoU) to comprehensively verify the performance of our method from multiple aspects.

\begin{table}[tb]
    \caption{\textbf{Performance comparison of four evaluation indicators for image size 336.} The best average performance under each metric is bolded.
  }
  \label{tab:main336}
    \centering
    \scalebox{0.84}{
    \renewcommand\arraystretch{1.3}
    \begin{tabular}{c|cccc|cccc|cccc}
    \toprule
        Image size=336 & \multicolumn{4}{c|}{CLIP} & \multicolumn{4}{c|}{SCLIP} & \multicolumn{4}{c}{CLIPtrase (Ours)} \\
        Evaluation & pAcc & mAcc & fwIoU & mIoU & pAcc & mAcc & fwIoU & mIoU  & pAcc & mAcc & fwIoU & mIoU \\ \midrule
        COCO-obj & 22.53 & 19.85 & 13.55 & 12.63 & 48.34 & 57.62 & 36.79 & 40.43 & 50.01 & 62.55 & 38.24 & 44.84 \\
        VOC21 & 43.16 & 45.97 & 31.39 & 17.31 & 79.41 & 82.66 & 68.64 & 51.79 & 79.93 & 85.24 & 69.10 & 53.04 \\ 
        PC60 & 21.84 & 21.25 & 11.20 & 8.91 & 49.95 & 51.45 & 35.88 & 29.00 & 53.21 & 56.43 & 38.76 & 30.79 \\ \hline
        COCO-stuff & 11.11 & 11.95 & 5.82 & 4.70 & 35.97 & 39.51 & 25.08 & 21.61 & 40.14 & 45.09 & 27.96 & 24.06 \\
        VOC20 & 57.12 & 57.92 & 42.60 & 41.06 & 87.03 & 90.97 & 78.48 & 79.12 & 89.51 & 91.77 & 82.15 & 81.20 \\ 
        PC59 & 24.53 & 21.81 & 13.42 & 9.82 & 56.50 & 52.49 & 43.08 & 32.59 & 60.15 & 57.47 & 46.64 & 34.92 \\ 
        PC459 & 16.94 & 3.94 & 9.50 & 1.88 & 42.07 & 18.43 & 33.73 & 8.97 & 45.77 & 20.72 & 36.67 & 10.11 \\
        ADE150 & 5.64 & 7.65 & 2.89 & 2.3 & 33.84 & 32.28 & 24.27 & 14.76 & 39.92 & 37.75 & 29.17 & 17.04\\ 
        ADEfull & 2.96 & 3.04 & 1.39 & 0.79 & 21.33 & 15.3 & 15.9 & 5.28 & 26.73 & 17.99 & 20.3 & 5.89\\ \hline
        AVG. & 22.87 & 21.49 & 14.64 & 11.04 & 50.49 & 48.97 & 40.21 & 31.51 & \textbf{53.93} & \textbf{52.78} & \textbf{43.22} & \textbf{33.54} \\
    \bottomrule
    \end{tabular}}
\end{table}

\subsection{More Experiment Results}
\label{b2}
We mainly reproduce the semantic segmentation of CLIP\cite{clip} and SCLIP\cite{sclip} on various datasets, and mainly compare the two models to analyze the effectiveness of our model in the training-free open vocabulary semantic segmentation.

Compared with the results in the main text, we mainly supplement the different 
image sizes of pAcc, mAcc, fwIoU and mIoU using CLIP, SCLIP and our own methods on each dataset in Table \ref{tab:main224} and \ref{tab:main336}, to prove the effectiveness of our method from a more comprehensive perspective.
Judging from the improvement of pAcc, our method distinguishes and clusters objects with different semantics in the image, rather than just focusing on the main object. Although some objects are uniformly regarded as background, we think this advantage will have greater potential in subsequent downstream tasks.

In addition, under these measurement standards, no matter with which resolution, our model is about 3\% higher than SCLIP, which can more comprehensively prove the effectiveness of our method.

\begin{table}[tb]
  \centering
  \caption{\textbf{Results on unsupervised semantic segmentation.} We use datasets that are consistent with the baselines, with the dataset suffix indicating the number of categories.}
  \setlength{\tabcolsep}{1mm}{
  \begin{tabular}{c|cccc|c}
    \toprule
   mIoU & \makecell[c]{PiCIE~\cite{picie}\\(CVPR'21)} & \makecell[c]{STEGO~\cite{eagle}\\(ICLR'22)} & \makecell[c]{HP~\cite{hp}\\(CVPR'23)}  & \makecell[c]{SmooSeg~\cite{smooseg}\\(NIPS'23)} & \textbf{Ours}\\
   \hline
   coco27 & 13.8 & 24.5 & 24.6 & 26.7 & \textbf{30.8} \\
   cityscape27 & 12.3 & 21.0 & 18.4 & 18.4 & \textbf{20.0} \\ \bottomrule
  \end{tabular}}
  \label{table:uss}
\end{table}

\subsection{More applications}
\label{b3}
In addition to the application mentioned in the main paper that involve combining our model with SAM, our model can be applied to many other areas as well.

\mypara{Combining with unsupervised semantic segmentation.}
Our CLIPtrase model, by recovering the internal local correlations within CLIP through self-correlation, enables CLIP to provide more semantic contextual details within images. This kind of information is invaluable for tasks that lack pixel-level annotations, such as semi-supervised and unsupervised semantic segmentation.

Taking unsupervised semantic segmentation tasks as an example, we combine the generalization capability of Clip with the region correlations recovered by our method and apply them to the unsupervised task. We compare our approach with SOTA models in Table \ref{table:uss} and find that our CLIPtrase plays a significant role in unsupervised semantic segmentation.

\mypara{Combining with training OVSS.}
Additionally, our method can also assist OVSS models that require training. For instance, for the SAN~\cite{san} model, we utilize the features and masks from CLIPtrase and perform mask average pooling (MAP) to initialize the query embedding in SAN.

\begin{figure}[tb!]
\centerline{\includegraphics[scale=0.2]{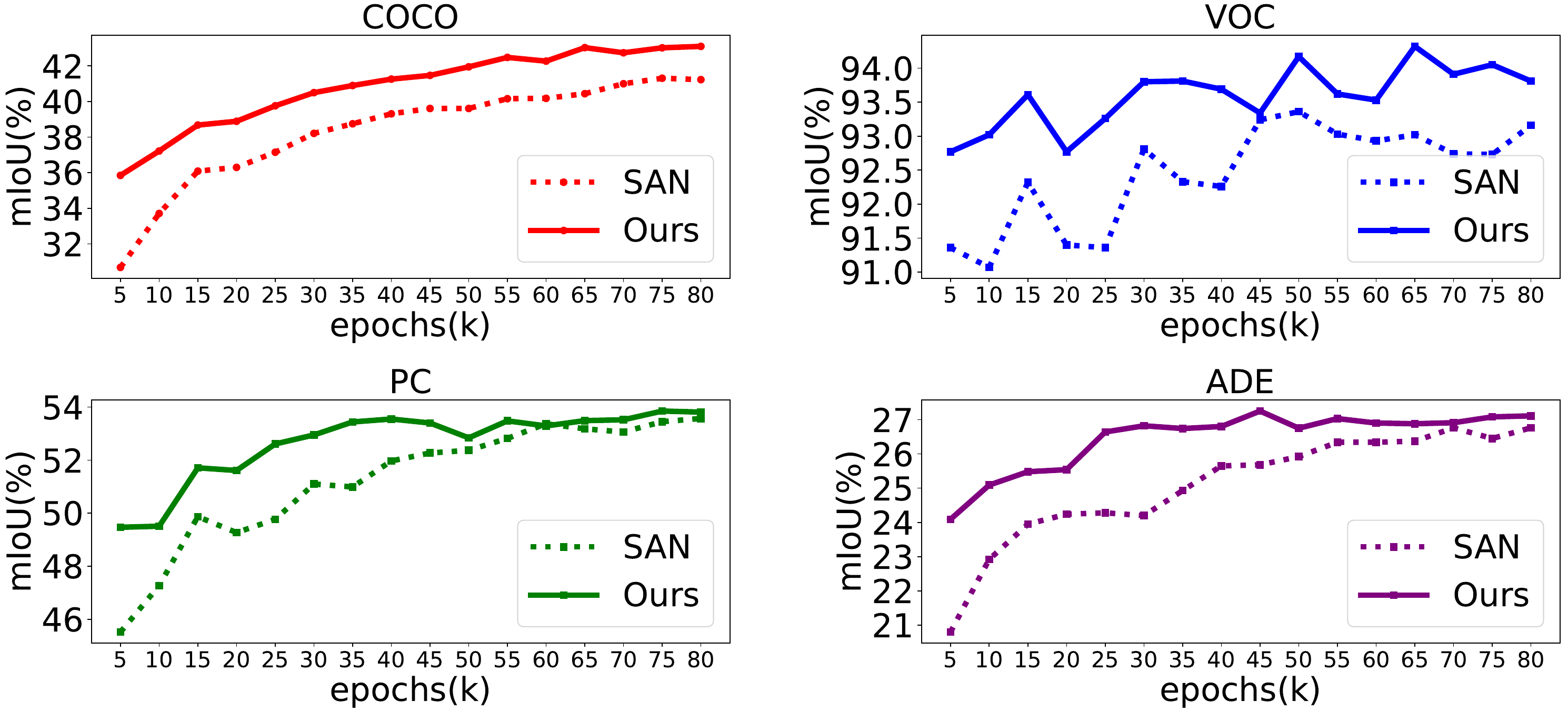}}
\caption{\textbf{The convergence of our approach combined with SAN.}}
\label{fig:san}
\end{figure}

Our method helps improve the training speed of the SAN model. SAN itself is a lightweight model, when combined with our approach, which shown in Figure \ref{fig:san}, it further reduces the training burden, resulting in a doubling of the convergence speed on specific datasets.

\section{Related Work}
\label{c}
\subsection{Contrastive Language-Image Pre-training}
\label{c1}
Contrastive Language-Image Pre-training (CLIP)\cite{clip} is a large multi-modal foundation model, which utilizes the contrastive training of aligning visual and text corresponding category features, greatly improves the generalization on unseen samples. Currently CLIP is widely used in Few-Shot/Zero-Shot Learning (FSL/ZSL)\cite{coop,cocoop,maple,rpo,clipes}, Prompt learning\cite{coop,cocoop,maple,rpo} and Out-of-Distribution (OoD)\cite{clipood} tasks. 
Later, researchers begin to apply CLIP to dense feature tasks\cite{icar,openseed,regionclip,proposalclip} such as semantic segmentation\cite{clipes,cgformer}.

Li et al.\cite{clipsurgery} elaborate on the inherent noise problem of CLIP and introduce it into the open vocabulary task from the perspective of explainability through self-attention improvement. 
Unlike pipelines that generally fine-tune pre-trained models on additional data sets, the CLIP encoder often needs to be frozen and cannot be fine-tuned because it needs to maintain alignment with the text feature space\cite{maskclip}. Therefore, researchers currently prefer to use clip directly as an encoder to obtain preliminary features to inherit its excellent generalization ability, and pay more attention to design sophisticated decoders\cite{zegformer,denseclip,catseg,clsclip,mvpseg} to refine the image-level features to adapt to dense feature tasks.

\subsection{Open-Vocabulary Semantic Segmentation}
\label{c2}
Open-vocabulary semantic segmentation extends segmentation~\cite{pfenet,hdmnet,lsae,semi,cuigeneral,gfsseg,reslt,yingcong,jiang_semi,pfenet++,pointcept2023} and refers to segment semantic regions via textual names or descriptions for the open world without any mask annotations.
Early works\cite{maskclip} verify the importance of modal alignment in CLIP, and common downstream fine-tuning may destroys its generalization ability. 
MaskCLIP\cite{maskclip} attempts to improve the Vision Transformer (ViT)\cite{vit} structure of CLIP to allow the model to obtain coarse feature localization, and combines transductive learning to improve performance.
CLIP-Surgery analyzes the difficulty of the current semantic segmentation task introduced by CLIP from the perspective of image-text noise, and made certain improvements to the model using the idea of self-attention.
SCLIP\cite{sclip} inherits the idea of self-attention of MaskCLIP and directly adapts the improved CLIP structure to the semantic segmentation task.

Both CLIP-Surgery and SCLIP utilize the idea of self-attention to improve CLIP, while only CLIP-Surgery mentions the noise problem caused by the open category of text.
None of them explore and analyze why CLIP lacks the semantic correlation between patches.
Our work complements this point that it is the global patch formed during the attention interaction between [CLS] token and patches that leads to this.

The above methods attempt to directly apply CLIP to semantic segmentation tasks. In the methods of training on additional segmentation datasets, CLIP tends to be used as an encoder. We roughly divide them into two ideas:

\mypara{Decoder-based.} Inspired by MaskFormer\cite{maskformer} and Mask2Former\cite{mask2former}, open vocabulary semantic segmentation widely use the pipeline of mask generation + mask classification. 
This method trains the refined features using a pixel decoder and utilizes an additional query decoder to aggregate the refined features at different positions. It employs query embedding to obtain masks for different objects, calculates the similarity between query embedding and texts, and then weights the query masks accordingly to ultimately yield boundaries and categories for each class.

Thanks to the excellent architecture of MaskFormer, the effect of the mask generation module masks great progress. 
However, The generalization performance of the above mask classification on unseen samples is always the bottleneck of this problem\cite{ovseg} due to the limitation of training scale. Therefore, some researchers combine it with CLIP to complete the classification of masks through the generalization of CLIP and improve the model performance\cite{pmaskclip,scan,deop,tang2024mind,tang2023consistency,tang2023source}.

For example, Xu et al.\cite{san} design the side network to generate masks in parallel with CLIP, and then use the masks as the attention bias to learn the corresponding [CLS] token for each mask and complete mask classification. Similar ideas include TCL\cite{tcl}, which enables CLIP simultaneous participation in the mask generation and classification stages by reusing the visual branch.

Another method with the idea of combining CLIP and masks is GroupViT\cite{groupvit,zeroseg}. It designs multiple group tokens, continuously aggregates them during the text guidance training process, and finally makes each group token cover the area of a specific object. However, compared with the above idea, the prediction results of this method often contain messy and tiny segmented areas. 

\mypara{Fine-tuning.} In addition, there are also methods that advocate direct fine-tuning of the CLIP\cite{pacl,pfenet,ovseg,yang2023improved,yang2023lidar,liu2023vida}, among which the typical method is OV-Seg\cite{ovseg}.
It believes that the classification of images after masks is may have the domain shift problem, so  OV-Seg use additional masked dataset to fine-tune the CLIP, adapt to the special need of mask classification.
MAFT\cite{maft} draws on the idea of SAN\cite{san} and fine-tunes the process of generating corresponding [CLS] tokens based on the attention bias formed by different masks, so that it can also achieve the purpose of improving mask classification performance.

There are also methods to directly use the contrast learning idea of CLIP to retrain the encoder at the patch or pixel level. For example, PACL\cite{pacl} refines this alignment to the patch level and improves semantic coherence issues.

Despite the decent results demonstrated by the two approaches, they
still possess their respective issues. For the decoder-based method, the CLIP’s features are obtained via a frozen CLIP model without specific adjustments to adapt them for semantic segmentation. Instead, this issue is addressed through the addition of an external decoder. On the other hand, although fine-tuning allows for a certain degree of adaptation of CLIP’s image-text aligned features, it carries the risk of overfitting to specific scenarios, e.g., the domain of the masked images used for fine-tuning, leading to a decline in performance. Therefore, we need to consider whether it is possible to optimize CLIP’s features without finetuning, in order to unearth more information that can aid in semantic segmentation. To answer this question, we start by investigating the correlations between the [CLS] token and the patch tokens.

\section{Visualization}
\label{d}
In this section, we mainly visualize the effects of several modules in the method in detail. 

\begin{figure}[H]
  \centering
  \includegraphics[height=18cm]{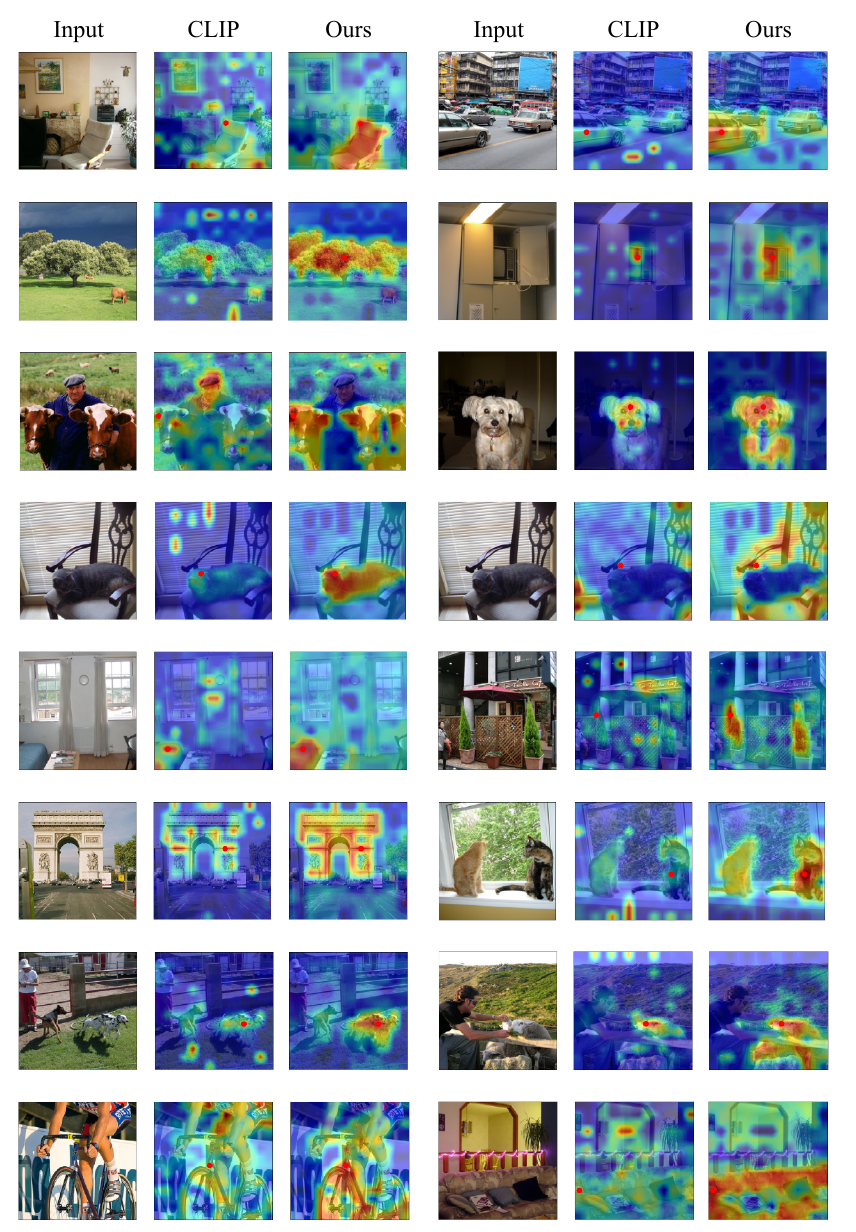}
  \caption{\textbf{More comparisons before and after semantic correlation recovery.} The red dot indicates the selected patch position.
  }
  \label{fig:visual1}
\end{figure}

Figure \ref{fig:visual1} illustrates the global patch problem we mentioned above, and the results improved with our semantic relevance recovery method.
We demonstrate this global patch phenomenon and the performance of our improved method in various image situations through richer visualizations. 

In the original CLIP, the response heat map of the area we select is completely inconsistent with the semantic area where it is located due to the global patch. However, with our semantic correlation recovery, this dilemma can be greatly improved, and surprising semantic correlations can also be observed among multiple objects with same semantics that are located at a considerable distance from each other.

\begin{figure}[H]
  \centering
  \includegraphics[height=18cm]{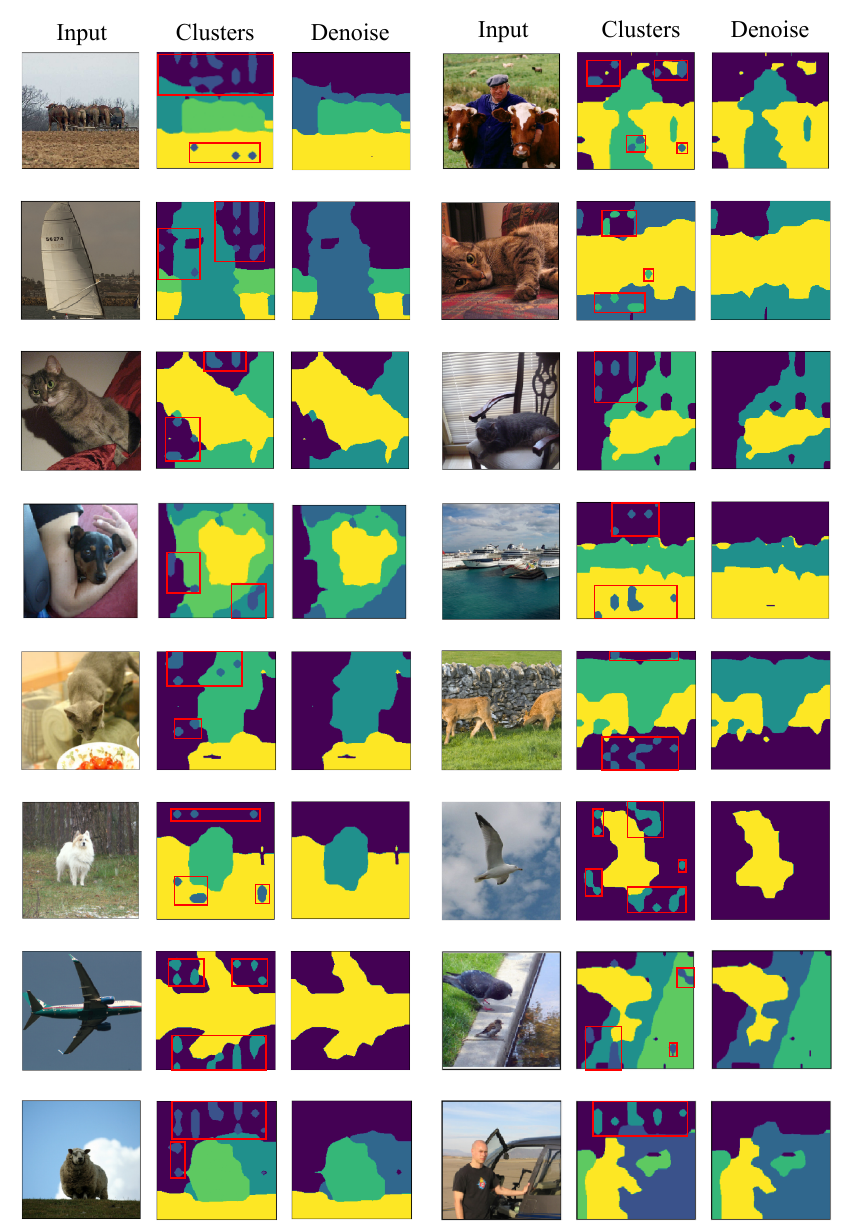}
  \caption{\textbf{Visualization of clustering results and denoising results in our method. }The red box represents the noise caused by the existing global patches.
  }
  \label{fig:visual2}
\end{figure}

Figure \ref{fig:visual2} shows the clustering effect of our model and the noise problem caused by the global patch. We perform denoising operations on this basis, and obtain masks with better qualities, finally finish the predictions of semantic segmentation. In the visualization, the result after denoising is actually the final semantic segmentation results, here we do not mark specific category labels on the images for simplicity.

\bibliographystyle{splncs04}
\bibliography{main}

\end{document}